\documentclass{article} 
\usepackage{iclr2027_conference,times}

\usepackage{amsmath,amsfonts,bm}

\def\eqref#1{equation~\ref{#1}}

\def\1{\bm{1}}

\DeclareMathAlphabet{\mathsfit}{\encodingdefault}{\sfdefault}{m}{sl}
\SetMathAlphabet{\mathsfit}{bold}{\encodingdefault}{\sfdefault}{bx}{n}

\usepackage{hyperref}
\usepackage{url}
\usepackage{graphicx}
\usepackage{caption}
\usepackage{float}
\usepackage{placeins}
\usepackage{booktabs}

\title{
Hyperbolic Inverse Reinforcement Learning
}

\author{Iulia Rusu \\
University of California San Diego \\
Salk Institute for Biological Studies \\
\texttt{irusu@ucsd.edu} \\
\And
Anoop Praturu \\
Causal Labs \\
\texttt{anoop@getcausal.ai} \\
\AND
Sergei Gepshtein \\
University of California San Diego \\
\texttt{sgepshtein@ucsd.edu} \\
\And
Tatyana O. Sharpee \\
University of California San Diego \\
Salk Institute for Biological Studies \\
\texttt{sharpee@salk.edu}
}

\iclrfinalcopy 
\begin{document}

\maketitle
\thispagestyle{fancy}
\lhead{Under review as a conference paper at ICLR 2027}

\begin{abstract}

Inverse reinforcement learning (IRL) seeks to infer the reward function underlying observed 
behavior, but its predictive accuracy and interpretability depend on how that reward is 
computed from the underlying features. While prior IRL reward functions have been implemented 
through a variety of techniques including linear weights, neural networks, or other function 
approximators, none make assumptions about the geometry of the underlying feature space. 
Inspired by geometric organization observed in neural circuits, we introduce Hyperbolic inverse
reinforcement learning (Hyperbolic IRL), which parameterizes reward as a function of Hyperbolic
distance between the feature vector and a learned parameter vector, as well as its Euclidean 
distance variant. We tested our methods against a range of linear and nonlinear IRL baselines 
under both deterministic and stochastic policies. Our Hyperbolic approach achieved comparable 
performance to state-of-the-art models on established benchmarks and excelled in the low data 
limit and across dense environments. Applied to human navigation among multiple exhibits in a 
museum room, Hyperbolic IRL reveals reward structure that depends not only on individual 
exhibit-distance features, but also on non-separable conjunctions between them. Together, 
these results demonstrate that a Hyperbolic reward parameterization can improve both the 
predictive accuracy and interpretability of IRL, providing a framework for studying how 
non-separable, conjunctive structure can begin to explain behavior.

\end{abstract}

\section{Introduction}

Inverse Reinforcement Learning (IRL) methods learn a reward function from expert demonstrations, formulated
within a Markov Decision Process (MDP) where states are represented through a feature function and reward
is recovered over these features rather than the states themselves. Learning a generalizable policy this
way is challenging: many reward functions can explain the demonstrations equally well, yet correspond to
policies that overfit and fail to generalize \citep{ng2000algorithms}. Thus, it is important to not only
explain the demonstrations, but to do so with an interpretable reward structure. Existing methods impose
such structure by fixing how reward is computed from features: classic approaches take reward to be a
linear combination of hand-selected features \citep{ng2000algorithms, abbeel2004apprenticeship}, while more
recent work replaces this linear combination with a more expressive function approximator such as a
Gaussian process \citep{levine2011nonlinear} or a neural network \citep{wulfmeier2015maximum}. Whether
linear or nonlinear, however, these approaches all assume the underlying feature space is flat.

As environments grow more complex, so does the dimension of the feature and reward space. A common
assumption in ML is that despite this high dimension, correlations among features let the data be
described by a lower-dimensional manifold. But the right geometry for that compression, not just its
dimension, depends on the structure of the data \citep{nickel2017poincare}. Assuming a flat manifold
when the true structure is curved can distort the compression. In particular, hyperbolic space is known to
pack substantially more structure into a small number of dimensions than a flat space of the same
dimension can: because its volume grows exponentially rather than polynomially with radius, it can
embed hierarchical or exponentially-branching structure at far lower distortion using far fewer
dimensions \citep{cannon1997hyperbolic, krioukov2010hyperbolic, nickel2017poincare}. Extending standard
machine learning methods to curved manifolds, however, is not straightforward.
Most methods, such as matrix factorization \citep{mnih2007probabilistic} and neural networks
\citep{vaswani2017attention}, measure similarity
between vectors using the dot product, which is a fundamentally Euclidean notion with no
natural analogue on a curved space \citep{cannon1997hyperbolic}.
Distance, by contrast, is well defined on any Riemannian manifold, flat or curved. 
\citet{praturu2025distance} show that replacing the dot product with a distance-based 
similarity measure not only makes this extension to curved geometries possible, but is itself 
a more expressive and better-generalizing choice than the dot product even when curvature
is held fixed, a pattern also found in other domains, e.g.\ neural population encoding
\citep{wu2026hyperbolic} and drug-target interaction prediction \citep{poleksic2023hyperbolic}.
We adopt the same distance-based approach for reward in IRL:
reward is computed as distance between the feature vector and a learned parameter vector, 
so that the same functional form can be instantiated in either a Euclidean or a Hyperbolic feature space 
without any change to its basic structure. Having multiple curvature options allows one to better fit to the true shape of the reward manifold.

Our distance-based geometric approach was also inspired by the Hyperbolic organization found in neural circuits \citep{zhang2023hippocampal, zhou2018hyperbolic}.
Meanwhile IRL itself is an increasingly common tool for analyzing behavior and decision-making in neuroscience \citep{ashwood2022dynamic, ke2025switching}, 
so a reward geometry grounded in the brain's own representational structure is a natural fit for this growing use case.
Despite this, Hyperbolic distance has not previously been used for IRL, although Hyperbolic geometry has
been applied broadly elsewhere in ML: to hierarchical embeddings \citep{nickel2017poincare}, neural network
architectures \citep{ganea2018hyperbolic}, graph representation learning \citep{chamberlain2017neural}, and
the geometry of complex networks \citep{krioukov2010hyperbolic}.

Our contribution is a framework for Distance-Based Inverse Reinforcement Learning, based on Maximum Entropy IRL \citep{ziebart2008maximum}.
We define the reward to be a function of distance between the feature vector and learned parameter vector in either Hyperbolic or Euclidean space.
We argue that the distance-based reward, and the Hyperbolic variant in particular, offers several advantages: the flexibility to select a curvature suited to the 
true geometry of the reward, a substantial reduction in the number of parameters needed to achieve competitive performance relative to more heavily parameterized 
function approximators, a closer biological relevance to the geometry observed in biological neural circuits, and an interpretable reward structure as a function of 
joint and separate feature components.

\section{Method: Distance-based Rewards}

\subsection{MDPs and Maximum Entropy IRL}
\label{sec:prelim}

We consider a Markov Decision Process $\mathcal{M} = \{\mathcal{S}, \mathcal{A}, \mathcal{T}, \gamma\}$, where $\mathcal{S}$ is the
state space, $\mathcal{A}$ is the action space, $\mathcal{T}: \mathcal{S} \times \mathcal{A} \times \mathcal{S} \to [0,1]$ is the transition
distribution, and $\gamma \in [0,1]$ is the discount factor. Inverse reinforcement learning (IRL) takes in observed expert demonstrations $\mathcal{D} = \{\tau_1, \dots, \tau_N\}$, with each consisting of state and action pairs $\tau_i = \{(s_1, a_1), (s_2, a_2), \dots\}$.
There is a reward value associated with each state $R(s_t)$. IRL tries to find the reward function $R$ responsible for the set of expert demonstrations.

While there are many IRL frameworks, including margin-based \citep{ratliff2006maximum}, Bayesian \citep{ramachandran2007bayesian},
and adversarial \citep{fu2018learning}, we use Maximum Entropy IRL (MaxEnt IRL) \citep{ziebart2008maximum}, which handles expert 
suboptimality and environmental stochasticity by modeling a full distribution over trajectories rather than 
assuming the expert acts optimally. In MaxEnt IRL, the state $s_t$ is described by a feature vector $\vec{\phi}(s_t)$. 
The framework states that among all trajectory distributions whose expected features match the expert's observed feature expectations, 
the one with maximum entropy is the most correct. This yields a distribution over
trajectories that is exponential in the accumulated reward:
\begin{equation}
    P(\tau \mid \theta) = \frac{1}{Z(\theta)} \exp\left( R(\tau) \right), \qquad
    R(\tau) = \sum_{t=1}^{T} R(s_t),
\end{equation}
where $Z(\theta) = \sum_{\tau} \exp(R(\tau))$ is the partition
function. $\theta$ are the parameters of the reward function. The reward function for a single state was originally written
as a function of the state's features $\vec{\phi}(s_t)$ and the learned $\vec\theta$ vector, with the assumption that reward
is linear in these features \citep{ziebart2008maximum}.
\begin{equation}
    R(s_t) = \vec{\phi}(s_t) \cdot \vec{\theta}
    \label{eq:linear-reward}
\end{equation}
Fitting for $\theta$ by maximum likelihood gives the general IRL loss
\begin{equation}
    \mathcal{L}(\theta) = -\frac{1}{N} \log P(\mathcal{D} \mid \theta) = \log Z(\theta) - \frac{1}{N} \sum_{i=1}^{N} R(\tau_i),
    \label{eq:general-loss}
\end{equation}
which is convex in $\theta$ whenever $R(\tau)$ is linear in $\theta$, and whose gradient is:
\begin{equation}
    \nabla_\theta \mathcal{L}(\theta) = \mu_\pi - \mu_E,
    \label{eq:general-grad}
\end{equation}
where $\mu_\pi = \mathbb{E}_{\tau \sim \pi_\theta}\left[ \nabla_\theta R(\tau) \right]$ denotes the expected reward gradient under the trajectory-visitation distribution of the policy,
and $\mu_E = \frac{1}{N}\sum_{i=1}^{N} \nabla_\theta R(\tau_i)$ denotes the corresponding empirical average over the expert demonstrations. 

What varies across our proposed models is the parametric form of $R$ in Equation~\ref{eq:linear-reward};
we introduce the notion of ``distance'' between a state's features $\vec{\phi}(s_t)$ and the reward
parameters $\vec\theta$.

\subsection{Euclidean Distance}
\label{sec:euclidean}

We take $R(s_t) = -\tfrac{1}{2} d^2$, the negative squared Euclidean distance between $\vec{\phi}(s_t)$ and $\vec\theta$,
which expands to a linear-in-features form
\begin{equation}
    R(s_t) = \vec{\phi}(s_t) \cdot \vec{\theta} - \tfrac{1}{2} \left( \|\vec{\phi}(s_t)\|^2 + \|\vec{\theta}\|^2 \right).
\end{equation}
Substituting into Equation~\ref{eq:general-loss} and dropping the constant $\|\vec\phi\|^2$ term gives
\begin{equation}
    \mathcal{L}(\theta) = \log Z(\theta) - \mathbb{E}_{\mu_E}\left[ \vec{\phi}(s_t) \cdot \vec\theta - \tfrac{1}{2}\|\vec\theta\|^2 \right],
\end{equation}
with gradient
\begin{equation}
    \nabla_\theta \mathcal{L}(\theta) = \mathbb{E}_{\mu_\pi}\left[ \vec\phi(s_t) - \vec\theta \right] - \mathbb{E}_{\mu_E}\left[ \vec\phi(s_t) - \vec\theta \right].
\end{equation}
The full step-by-step derivation is given in Appendix~\ref{app:euclidean}.

\subsection{Hyperbolic Distance}
\label{sec:hyperbolic}

We keep $R(s_t) = -\tfrac12 d^2$, but now measure $d$ in Hyperbolic space. We use the Lorentz model of
Hyperbolic space $\mathbb{H}^d$: the upper sheet of a two-sheeted hyperboloid embedded in $(d+1)$-dimensional
Minkowski space, equipped with the Minkowski inner product $\langle \vec u, \vec v \rangle_{\mathcal{M}} = -u_0
v_0 + \sum_{n=1}^{d} u_n v_n$. A point $\vec x \in \mathbb{R}^{d+1}$ lies on $\mathbb{H}^d$ exactly when
$\langle \vec x, \vec x \rangle_{\mathcal M} = -1$ and $x_0 > 0$.

Our feature vector $\vec\phi(s_t)$ and the learned parameter vector $\vec\theta$ are both $d$-dimensional,
one dimension short of the ambient space $\mathbb{R}^{d+1}$ that $\mathbb{H}^d$ lives in. To put them into Hyperbolic space we add an additional component to each. Given a $d$-dimensional vector $\vec v$, we set $v_0 =
\sqrt{1+\|\vec v\|^2}$ and take $(v_0, \vec v) \in \mathbb{R}^{d+1}$ as the corresponding point on
$\mathbb{H}^d$, since $\langle (v_0,\vec v), (v_0,\vec v)\rangle_{\mathcal M} = -v_0^2+\|\vec v\|^2 = -1$ by
construction. Applying this to both $\vec\phi(s_t)$ and $\vec\theta$ gives $\phi_0(s_t) = \sqrt{1+\|\vec\phi(s_t)\|^2}$
and $\theta_0 = \sqrt{1+\|\vec\theta\|^2}$, so that $\theta$ now sits on the hyperboloid in this
$(d+1)$-dimensional space, alongside $\vec\phi(s_t)$.

The distance between the two points on the hyperboloid is
\begin{equation}
    d = \operatorname{arcosh}\left( -\langle \vec\theta, \vec\phi(s_t) \rangle_{\mathcal M} \right),
\end{equation}
so that
\begin{equation}
    R(s_t) = -\tfrac12 \left[ \operatorname{arcosh}\left( -\langle \vec\theta, \vec\phi(s_t) \rangle_{\mathcal M} \right) \right]^2 .
\end{equation}
This can now be optimized directly in hyperbolic space using Riemannian gradient descent, the details of which we leave to Appendix~\ref{app:hyperbolic}.

\subsection{Generalizing to Arbitrary Curvature}
\label{sec:curvature}

The formulation above fixes the hyperboloid to unit radius, $\langle \vec\theta, \vec\theta \rangle_{\mathcal M} =
-1$, corresponding to constant curvature $-1$. We generalize to a hyperboloid of radius $R$ (constant curvature
$-1/R^2$) by rescaling the defining constraint to $\langle \vec\theta, \vec\theta \rangle_{\mathcal M} = -R^2$,
i.e.\ $\theta_0 = \sqrt{R^2 + \|\vec\theta_{1:d}\|^2}$, so that the distance between two points becomes
\begin{equation}
    d_R\!\left(\vec\theta, \vec\phi(s_t)\right) = R \operatorname{arcosh}\!\left( -\frac{\langle \vec\theta, \vec\phi(s_t) \rangle_{\mathcal M}}{R^2} \right),
\end{equation}
with reward $R(s_t) = -\tfrac12 d_R^2$ as before. We denote this constant curvature $\kappa = -1/R^2$, and
refer to a given radius-$R$ model as Hyperbolic ($\kappa=-1/R^2$) throughout the rest of the paper. Riemannian
gradient descent generalizes in a similar way, with the gradient correction and exponential map each picking
up a factor of $R$; full details in Appendix~\ref{app:riemannian}. This lets us treat curvature as a
hyperparameter and compare reward geometries across a range of curvatures rather than fixing it to $-1$.

\section{Results}
\label{headings}

We tested the performance of our distance-based reward models using the Objectworld benchmark \citep{levine2011nonlinear}, as well as
a real world benchmark.

\subsection{Objectworld: Ground truth Reward Recovery}

\begin{figure}[htb]
\centering
\includegraphics[width=0.8\linewidth]{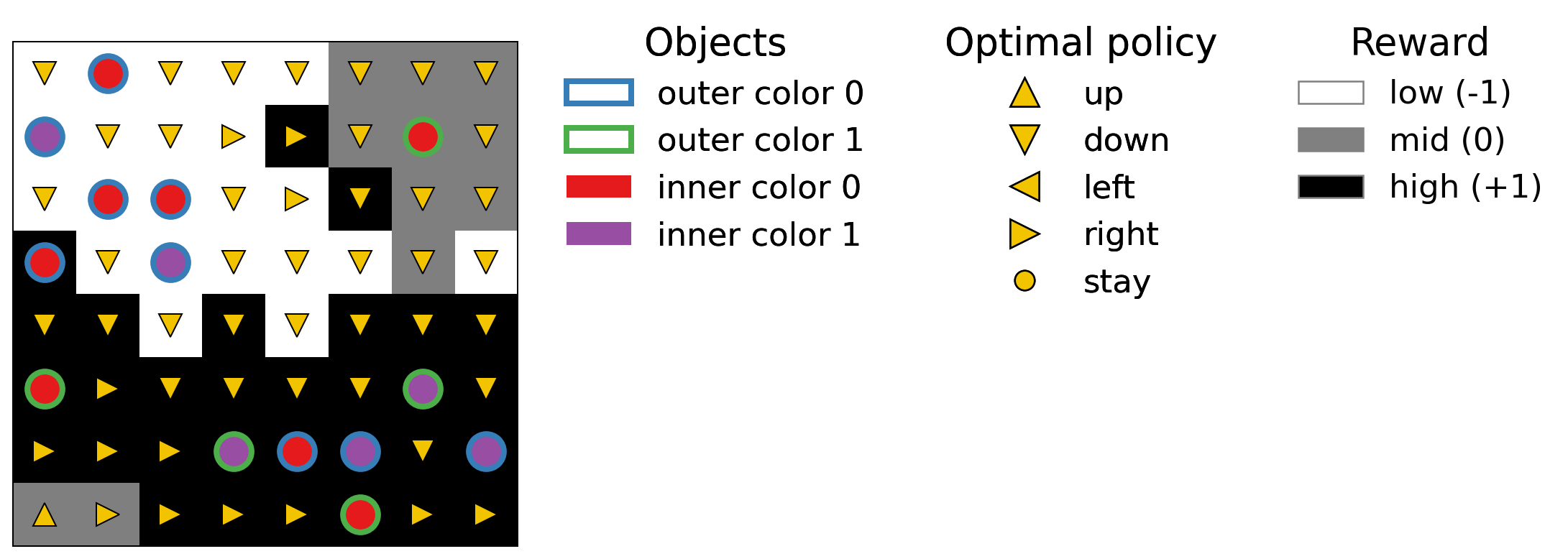}
\caption{Example $8\times8$ Objectworld environment with 14 objects, $C=2$ colors. True reward is
$+1$ (black) within 3 cells of an outer-color-0 object AND within 2 cells of an outer-color-1
object, $-1$ (white) within 3 cells of an outer-color-0 object alone, and $0$ (mid gray) otherwise.
Yellow arrows show the optimal policy under the reward.}
\label{fig:objectworld_setup}
\end{figure}

Objectworld is an $N\times N$ grid on which $n_{\text{objects}}$ objects are placed on random,
non-overlapping cells. Each object is independently assigned an inner and an outer color drawn uniformly from $C$ colors; every
color contributes a feature component to the state description, but only two of the $C$ outer colors are ever relevant to
determining the reward, with every other color, including all inner colors, acting as a distractor feature that plays no role
in the reward function.

Each grid cell is represented by a $2C$-dimensional feature vector: the Euclidean distance from the cell to the nearest object of each of the $C$ outer colors,
followed by the distance to the nearest object of each of the $C$ inner colors (a color with no placed object gets a constant distance larger than any true in-grid distance, 
instead of $\infty$). The ground truth reward is a nonlinear function of only two of these $2C$ features, the distance to the nearest outer-color-0 and outer-color-1 object: 
$+1$ near both, $-1$ near outer-color-0 alone, $0$ otherwise. There is built-in stochasticity as selection of action has a $30\%$ chance of being selected randomly, the other $70\%$
of time, the action is the intended action. We illustrate a small version of the grid in Figure 1.

We tested our 3 distance-based reward functions: Euclidean, Hyperbolic ($\kappa=-1$) and Hyperbolic ($\kappa=-4$). We compare them to 
classical Maximum Entropy Inverse Reinforcement Learning (MaxEnt IRL) \citep{ziebart2008maximum} and DeepIRL \citep{wulfmeier2015maximum}.
We compare the performance of these models using the Expected Value Difference (EVD) metric, which measures how suboptimal the policy is under
the true reward. To do this, we use the learned reward from each respective method and compute the optimal deterministic policy using $\arg\max_a Q(s,a)$ under that reward.
We then measure the expected sum of discounted rewards under the policy. We also evaluate under a stochastic policy, induced by 
soft value iteration on each learned reward, $\pi(a\mid s) \propto \exp(Q_{\text{soft}}(s,a))$. 

\begin{figure}[htb]
\begin{center}
\includegraphics[width=\linewidth]{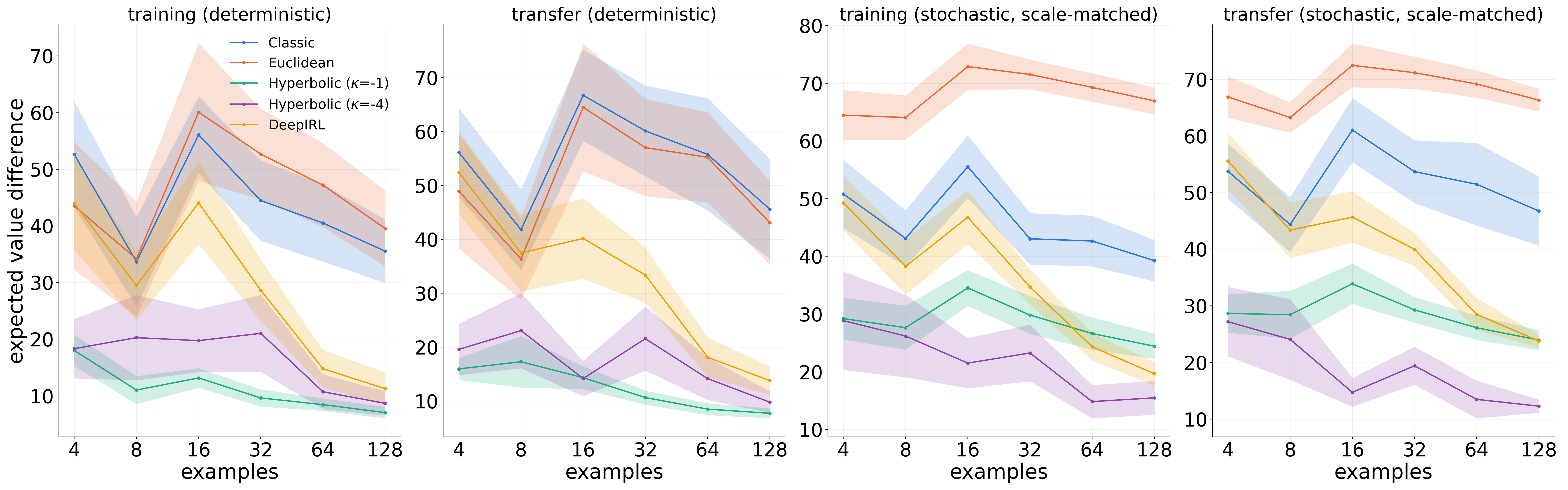}
\end{center}
\caption{Expected value difference vs. number of demonstration examples on $32\times32$
Objectworld with 50 objects, Classic, Euclidean, Hyperbolic ($\kappa=-1$), Hyperbolic ($\kappa=-4$),
and a neural network reward (16 repeats per condition). Hyperbolic Distance with negative curvature of $\kappa=-4$
had the best performance with increased sample count.}
\label{fig:EVD Across Number of Training Examples}
\end{figure}

We report the EVD comparison as the number of examples from the ground truth policy used for training increases, in Figure 2.
We also evaluate the policies on a transfer to 4 held-out environments, where
we change the location of objects in the grid, but not the size or the number of objects. We observe
that both Hyperbolic models outperform the others, with Hyperbolic ($\kappa=-4$) showing the best performance with increased examples in training, across
both stochastic and deterministic policies and in transfer environments. In Figure 3, we see that both Hyperbolic
variants lead in performance, with increasing object density in the grid.

\begin{figure}[htb]
\begin{center}
\includegraphics[width=\linewidth]{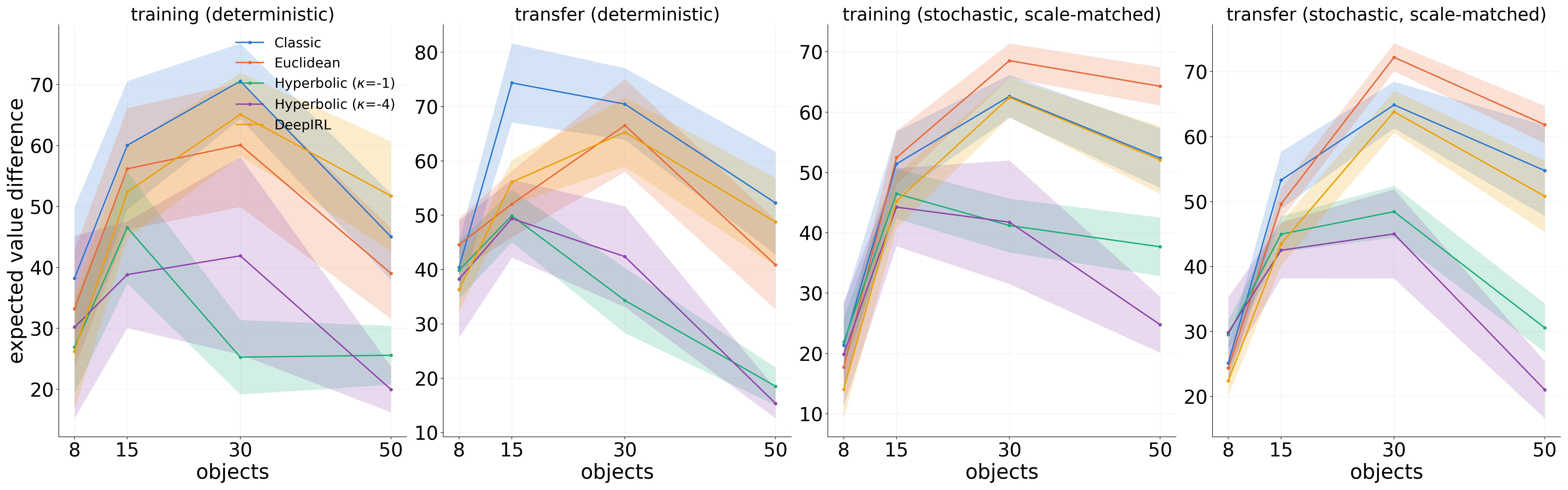}
\end{center}
\caption{Expected value difference vs. number of objects (8, 15, 30, and 50 objects, corresponding
to approximately $0.8\%$, $1.5\%$, $2.9\%$, and $4.9\%$ grid-cell density) on $32\times32$ Objectworld,
Classic, Euclidean, Hyperbolic ($\kappa=-1$), Hyperbolic ($\kappa=-4$), and DeepIRL. Hyperbolic Distance 
with curvature of $\kappa=-4$ showed the best performance out of all, with increased object density.
}
\label{fig:EVD Across Object Density}
\end{figure}

\FloatBarrier
\subsection{Museum}

\begin{figure}[htb]
\centering
\includegraphics[width=\linewidth]{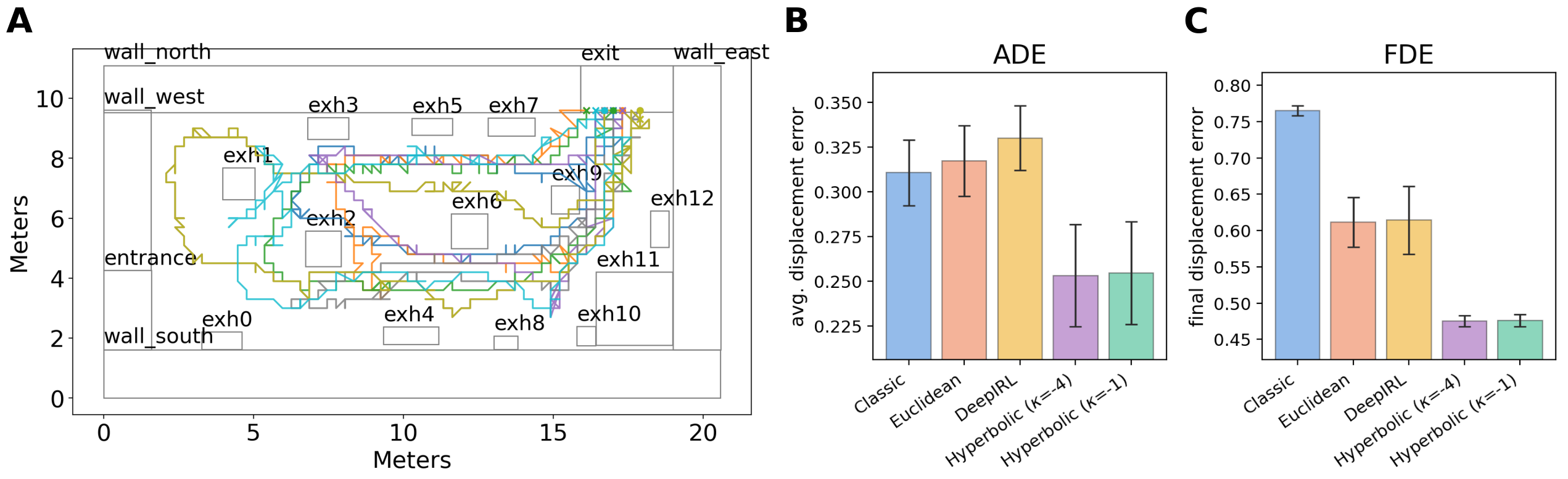}
\caption{A: Floorplan of LACMA ceramics gallery's 13 exhibits, with example walking trajectories colored by individual.
B, C: Average Displacement Error and Final Displacement Error of models trained on museum trajectories, error
bars showing standard error of the mean across the 4 held-out test trajectories.
Hyperbolic ($\kappa=-4$) is best at modeling the average trajectory and its starting and final locations.}
\label{fig:Museum Floorplan and Metrics}
\end{figure}

We test model performance on human behavioral data from HUMEX, a dataset of visitor movement in exhibition
spaces. The data used here were collected in a ceramics gallery at the Los Angeles County Museum of Art
(LACMA), where ten unobtrusive cameras mounted along the upper perimeter walls recorded visitor movement
during museum opening hours over several months. The recordings were processed to recover continuous
two-dimensional visitor trajectories in a common ground-plane coordinate system. The resulting trajectory
data are publicly available through the HUMEX repository on OSF \citep{gepshtein2026humex} (\url{https://osf.io/7f5du/}).
Figure~\ref{fig:Museum Floorplan and Metrics}A shows example trajectories from the dataset.
We discretized the floorplan into grid cells of $0.3m$ per side. The action space consisted
of five actions: stay, up, down, left, and right.
In the feature vector, the first 13 dimensions were the normalized Euclidean distance from the agent to each
respective exhibit; the last two were the agent's own normalized x and y position in the space. 
We trained on 18 of a total of 26 recorded trajectories, holding out 4 for validation and 4 for testing.

We quantified policy recovery using Average Displacement Error (ADE) and Final Displacement Error (FDE) 
which are standard metrics from the human trajectory-forecasting literature \citep{alahi2016social, xu2018collision}.
ADE measures, for each true trajectory, the mean Euclidean distance to 10 independent rollouts of the 
learned policy, each starting from the trajectory's own initial state. We averaged ADE from each true trajectory to get a final score. 
For FDE we do the same, but only use the final time point. The FDE is important in this specific scenario as all the trajectories in the analysis return to 
their initial location at the exit of the museum. Together these two measurements test how well the methods
recapitulate global patterns in the data.

In Figure~\ref{fig:Museum Floorplan and Metrics}B-C we note that Hyperbolic $\kappa=-4$ had the best ADE and
FDE performance out of all the models. Notably, it achieves this with 15 reward-function parameters,
compared to 1,601 for DeepIRL. 

\subsection{Feature Coupling}
\label{sec:feature-coupling}

\begin{figure}[htb]
\begin{center}
\includegraphics[width=\linewidth]{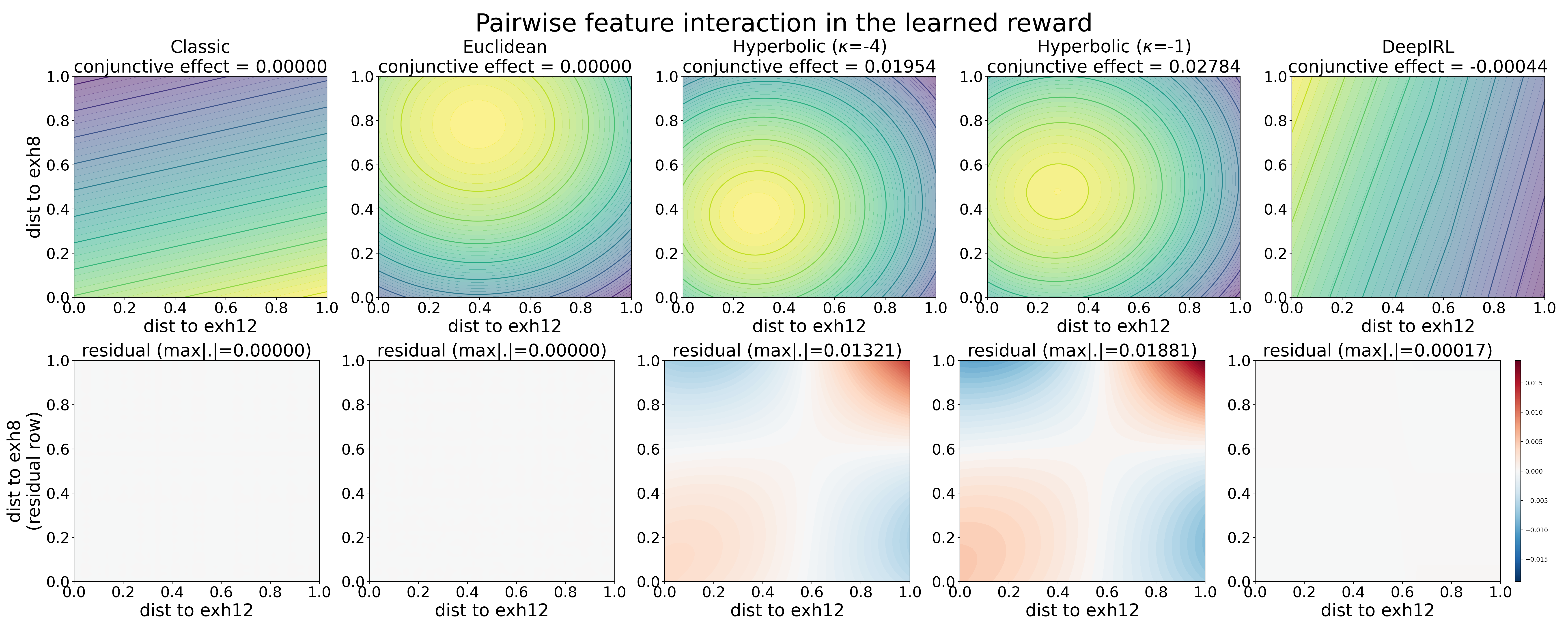}
\end{center}
\caption{Reward as a function of two feature dimensions (distance to exhibit 12 and distance to
exhibit 8), with the remaining 13 feature dimensions held at their median values, for each fitted
reward model. Top row: raw reward surface, as a function of two features. Bottom row: residual after removing the best additive
(two-way ANOVA main-effects) approximation of the surface.}
\label{fig:Feature Component Interaction}
\end{figure}

To test whether each reward model captures genuine interactions between feature components, rather than
treating every feature component as an independent additive contribution to reward, we probed the fitted
reward function directly. For a pair of exhibit-distance
feature components, we hold every other feature at its median value
and sweep $\phi_i,\phi_j$ over $[0,1]$, plotting the resulting reward surface $R(\theta,\phi)$
in Figure 5. 

We measured interaction using a two-way ANOVA-style fit: $R(\phi_i,\phi_j) \approx c +
\text{row-effect}(\phi_i) + \text{col-effect}(\phi_j)$, where the row/column effects are the
deviations of each row/column mean from the overall mean $c$. This additive approximation is
what a reward model would produce if $\phi_i$ and $\phi_j$ contributed to the reward
independently of one another; the residual, $R - (c+\text{row-effect}+\text{col-effect})$
(Figure~\ref{fig:Feature Component Interaction}, bottom row), is the part of the reward
surface that no additive combination of main effects can explain. A near-zero residual
indicates the two features act independently under that reward model, while visible structure in the
residual indicates a coupling between them. We saw that the Hyperbolic models were the only ones exhibiting this
non-flat residual, and therefore were experiencing non-separable feature coupling.

To summarize this non-additivity as a single number per feature pair, we also compute the mixed second
difference of the reward surface (full definition in Appendix~\ref{app:mixed-second-difference}). For ease
of reference, and because of its similarity to the concept of conjunctive coding in neuroscience, we will
refer to this quantity throughout the rest of the paper as the \emph{conjunctive effect}.

\begin{figure}[htb]
\centering
\begin{minipage}[c]{0.42\linewidth}
\centering
\includegraphics[width=\linewidth]{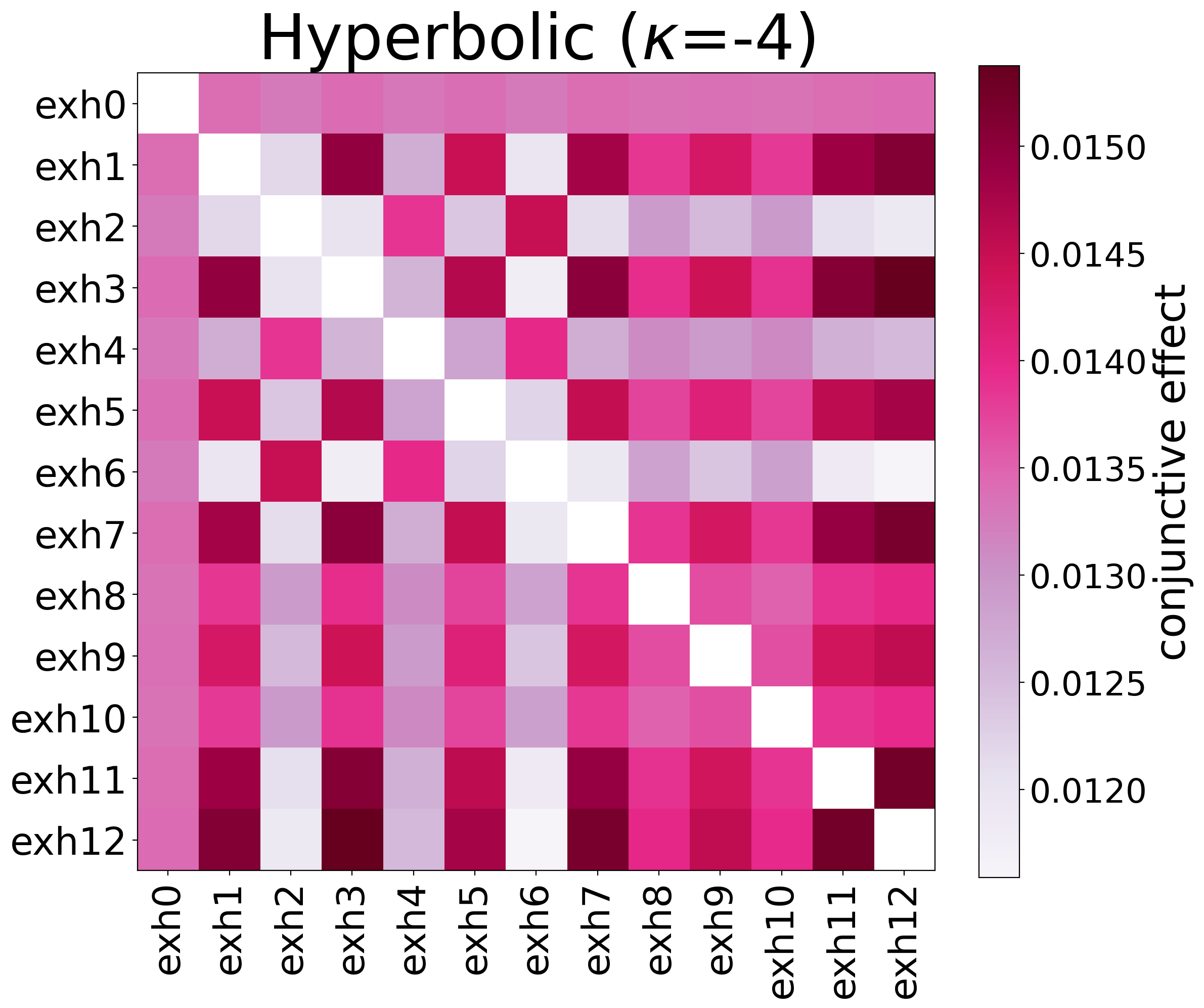}
\end{minipage}
\hfill
\begin{minipage}[c]{0.55\linewidth}
\centering
\includegraphics[width=\linewidth]{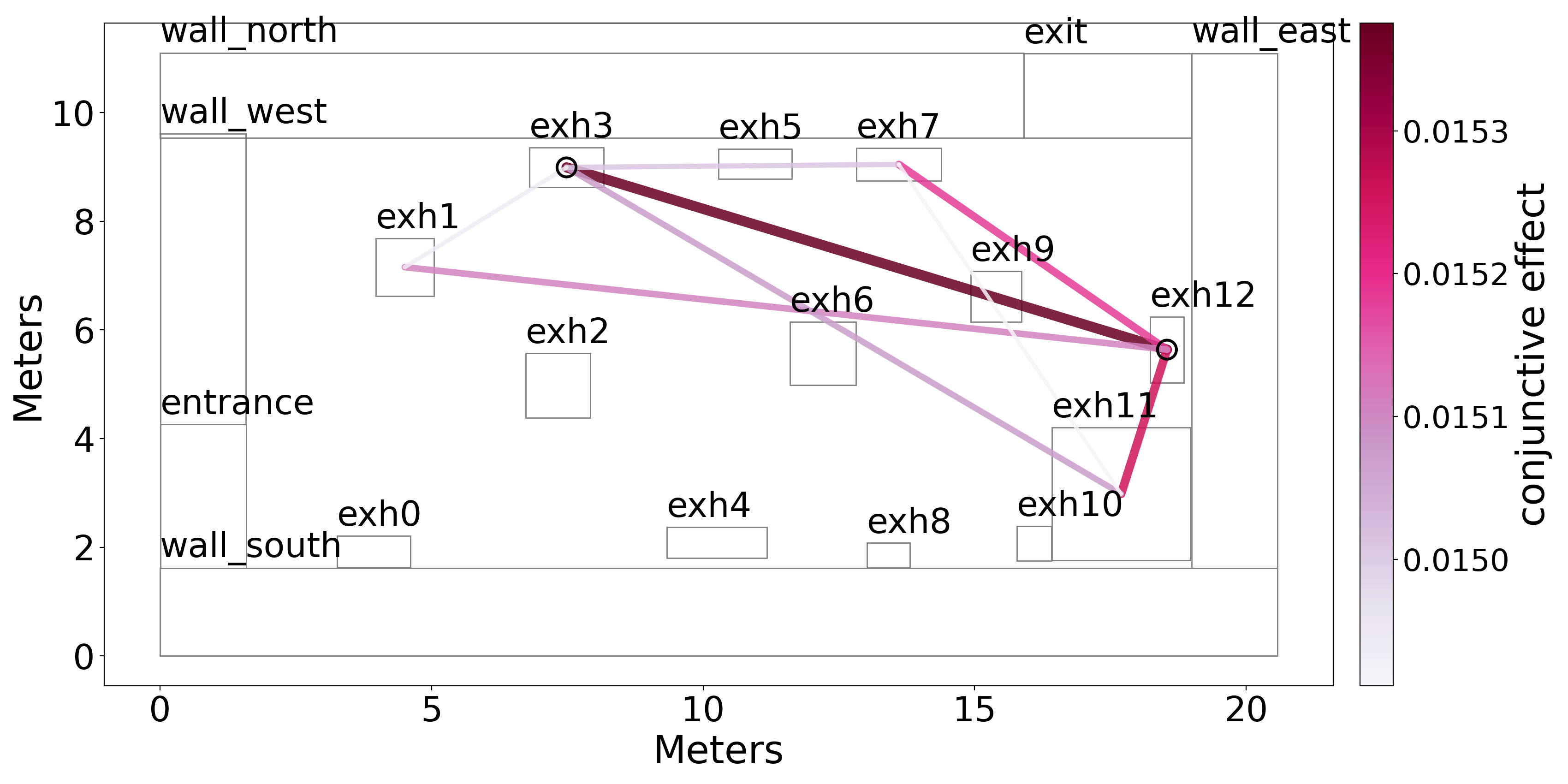}
\end{minipage}
\caption{Pairwise exhibit-distance interaction for the Hyperbolic ($\kappa=-4$) model. Left: the same
conjunctive effect measure as Figure~\ref{fig:Feature Component Interaction}, computed for all exhibit pairs. Right: the 8 strongest
conjunctive effects drawn on the museum floorplan, line width and color scaled to the conjunctive effect; exhibit 3 and exhibit 12
(circled) are hubs that every one of the 8 strongest pairs routes through.}
\label{fig:Exhibit Interactions}
\end{figure}

In Figure~\ref{fig:Exhibit Interactions} we analyze the conjunctive effect of the Hyperbolic ($\kappa=-4$) model, the geometry
with the best FDE and ADE performance on the museum dataset. We repeated the same conjunctive-effect probe used above for a single pair of exhibits, for every
pair of exhibit-distance feature components, shown as the matrix in Figure~\ref{fig:Museum Floorplan and Metrics}A.
Probing for the strongest conjunctive effects, we saw clusters around exhibit 12 and exhibit 3. Meanwhile the weakest all involved exhibit 6.

\begin{figure}[htb]
\centering
\includegraphics[width=\linewidth]{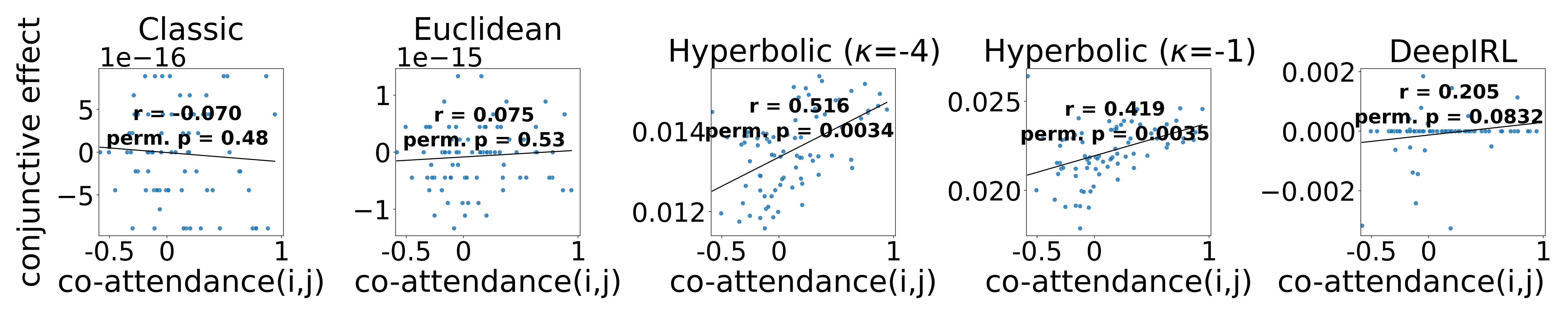}
\caption{Fitted feature interaction vs. real exhibit co-attendance, per geometry, for all 78 exhibit
pairs (allobjects), with a fitted line, Pearson $r$, and permutation-corrected $p$-value (10,000
shuffles of exhibit labels) shown per panel. Classic and Euclidean show no
relationship; both Hyperbolic variants shown here have a significant positive relationship (full
naive-vs-corrected comparison and a distance-controlled check in Table~\ref{tab:mantel},
Appendix~\ref{app:mantel-test}).}
\label{fig:CoAttendance Scatter}
\end{figure}

To test whether this fitted coupling of the conjunctive effect reflects genuine visitor behavior rather than an artifact of
the reward parameterization, we construct a purely behavioral measure, $\mathrm{co\_attendance}(i,j)$,
that does not use the fitted reward at all. For each visitor, we find how close they came to each
exhibit at any point during their visit; for a given pair of exhibits $i,j$, $\mathrm{co\_attendance}(i,j)$
is then the correlation, across all visitors, between how close they got to exhibit $i$ and how close
they got to exhibit $j$. A positive value indicates that visitors who sought out one exhibit tended to also seek
out the other. Figure~\ref{fig:CoAttendance Scatter}
plots, for each geometry, the fitted conjunctive effect for all 78 exhibit pairs against the real
$\mathrm{co\_attendance}$ value for that pair, with a fitted line, Pearson $r$, and permutation-corrected
$p$-value per panel. Because the 78 pairs are built from only 13 exhibits and are therefore not
independent samples, we assess significance with a permutation test rather than the naive $p$-value.
The full methodology, naive-vs-corrected comparison, and a distance-controlled partial-correlation check are
reported in Table~\ref{tab:mantel} (Appendix~\ref{app:mantel-test}). All three hyperbolic curvatures show
a significant positive relationship; Classic and Euclidean show essentially none, and this pattern is
unchanged after controlling for physical exhibit-to-exhibit distance, ruling out mere proximity as an
explanation (Appendix~\ref{app:mantel-test}).

\FloatBarrier

\FloatBarrier

\section{Discussion and Future Work}

In this work we introduce Hyperbolic IRL, a novel reward parameterization that improves both predictive
accuracy and interpretability while maintaining a low number of parameters, excelling especially in
low-data and high-density settings. Applied to museum visitor trajectories, we found that the inferred
reward contains a conjunctive effect between exhibit-distance features that corresponds to co-attendance
in the observed behavior, connecting to the neuroscience literature on conjunctive coding.

We offer two possible reasons for the Hyperbolic model's advantage over other models. Firstly, Hyperbolic space may 
be intrinsically better suited to representing densely distributed reward-relevant objects.
It is a classical fact that the volume of a ball in Hyperbolic space grows exponentially with
its radius, rather than polynomially as in Euclidean space \citep{cannon1997hyperbolic}, a property
already exploited to justify Hyperbolic geometry for hierarchical and scale-free structures
\citep{krioukov2010hyperbolic, nickel2017poincare}. \citet{praturu2025distance} note that this lets
points be packed densely around a reference location while remaining well-separated under the metric
the model actually uses.
In our setting, this means each object's zone of high reward can stay sharply localized 
around it without overlapping neighboring objects' zones, even as those objects become more numerous.
This offers an explanation for its advantage specifically as object density increases 
(Figure 3). 
Secondly, the Hyperbolic reward's coupling between feature components offers a complementary 
explanation. Section~\ref{sec:feature-coupling} showed that the Hyperbolic distance method captures a genuine 
conjunctive effect between exhibit-distance features, and that this conjunctive effect tracks real visitor behavior, 
unlike the additive Classic, Euclidean, and DeepIRL rewards. 
The reward's joint dependence on multiple features at once is unique to our model, but not a unique concept.

Indeed, selectivity for combinations of stimulus features, rather than single features in isolation, is a 
well-established property of sensory coding: classic characterizations of complex cells in visual cortex
and more general spike-triggered analyses both show that sensory neurons are typically best described by
several jointly-relevant stimulus features nonlinearly combined, rather than by one feature alone
{\NoHyper\citep{adelson1985spatiotemporal, rust2005spatiotemporal, schwartz2006spike, bialek2005features,
sharpee2013computational}\endNoHyper}. The conjunctive structure we find in the museum reward
function is consistent with this broader pattern of feature-combination sensitivity in biological sensory 
systems. It may prove to be an advantageous tool for behavioral neuroscience research. 

There has been a body of neuroscience research that focuses on studying decision-making and behavior using
IRL \citep{ashwood2022dynamic, ke2025switching, schultheis2021inverse, yamaguchi2018identification, kwon2020inverse, reddy2022reinforcement}.
However, much of the field remains bounded to studying behavior in artificial, simplified settings, due to
the relative ease of applying analytical tools in these more tractable contexts \citep{ashwood2022dynamic,
carandini2013probing}. We argue that the Hyperbolic method's unique advantages in performance in dense-object settings
with multiple reward sources and an underlying nonlinear reward structure make it a good candidate over
other, more high-parameter methods such as neural-network-based models. We reported each model's parameter count
in the Results section: 1,601 for DeepIRL versus 15 for every other model. A formal model-complexity comparison
(e.g., BIC) would require $N \gg K$ data points relative to parameters, which does not hold here given DeepIRL's 
parameter count relative to our training set size, so we do not report one. Nonetheless, the Hyperbolic model's 
substantially lower parameter count and its capacity to perform well in the low-data limit, as demonstrated 
in Figure 2, make it a good candidate for behavioral research, which is often constrained by limited 
sample sizes \citep{roy2021extracting, wilson2019ten}. A concrete example of a good use case for this method is 
foraging: the study of how animals search for and exploit resources, weighing decisions such as when to leave a depleting patch for a
potentially better one elsewhere \citep{stephens1986foraging, mobbs2018foraging, dalbello2021inversion, calhoun2015neural}. Arguably, the museum
dataset we presented is itself a type of foraging task, with visitors searching among and choosing between
multiple exhibits (reward sources) distributed through space. 

In future work, we will modify our model to learn the curvature of the Hyperbolic space directly via a Bayesian approach,
as was done for hyperbolic data embedding by \citet{praturu2024adaptive}, rather than testing over discrete sets of values. Another important direction will be testing whether
the conjunctive selectivity observed here generalizes to other real-world behavioral datasets, and, where paired neural 
recordings are available, whether it aligns with conjunctive coding in recorded neural activity.

\subsection*{AI use statement}

In this work, we used generative AI tools for refining our literature search, grammar help, and writing plotting code to generate scientific figures.
We have reviewed all AI-assisted work. We checked LLM-generated research ideas for potential plagiarism through a manual
literature survey and verified that the plotting of figures was done correctly.
We have not used generative AI tools to generate synthetic data sets, help develop theoretical models or conceptual frameworks, 
formulate mathematical claims, provide critical ingredients for proving mathematical claims, 
propose or refine hypotheses, design or provide feedback on research  methodology or experiments, implement methods,
clean and reformat datasets, support qualitative and thematic data analysis, and interpret results. 
The rest of the required disclosure tasks such as assistance in the writing of proofs, and assistance with translation are not applicable to this work.
We take responsibility for the final content of this work,
including text, claims or artifacts produced with the aid of generative AI.

\subsection*{Ethics statement}

This work uses human visitor trajectory data from HUMEX, a publicly available dataset
\citep{gepshtein2026humex} (\url{https://osf.io/7f5du/}) collected via unobtrusive overhead cameras in a public gallery at the
Los Angeles County Museum of Art. We only had access to de-identified, two-dimensional visitor
trajectories; no raw video, individual identities, or other personally identifiable information was
available to us. The Salk Institute Institutional Review Board granted exemption approval for this
project on January 7, 2022, under Common Rule 45 CFR 46.104(d)(2).
We do not foresee other ethical concerns regarding methodology, dual use, or fairness arising from
this work.






\raggedbottom
\bibliography{iclr2027_conference}
\bibliographystyle{iclr2027_conference}

\appendix
\section{Appendix}

\subsection{Euclidean Distance}
\label{app:euclidean}
We choose the following reward function $R(s_{t}) = -\frac{1}{2}d^2$ to use as a metric for the Euclidean distance between the feature vector $\vec{\phi}(s_{t})$ and
the reward parameters $\vec\theta$. The reward function is defined as follows: $R(s_{t}) = \vec{\phi}(s_t) \cdot \vec{\theta}$.
We wish to find the parameters of the reward function that maximize the probability of seeing the demonstrations, $P(\mathcal{D}|\theta)$.
We derive the loss function for the Euclidean distance reward model as follows:
\begin{align}
    \mathcal{L}(\theta) &= -\frac{1}{N} \log{P(\mathcal{D}|\theta)} \\
         &= -\frac{1}{N} \log \prod_{i=1}^{N} P(\tau_i|\theta) \\
         &= -\frac{1}{N} \sum_{i=1}^{N} \log P(\tau_i|\theta) \\
         &= -\frac{1}{N} \sum_{i=1}^{N} \log \frac{e^{R(\tau_i)}}{Z(\theta)} \\
         &= -\frac{1}{N} \sum_{i=1}^{N} \log( e^{R(\tau_i)}) + \frac{1}{N} \sum_{i=1}^{N} \log Z(\theta) \\
         &= \frac{1}{N} \sum_{i=1}^{N}\log Z(\theta) -\frac{1}{N}\sum_{i=1}^{N} \log( e^{R(\tau_i)}) \\
         &= \frac{1}{N} \sum_{i=1}^{N} \log Z(\theta) -\frac{1}{N} \sum_{i=1}^{N} R(\tau_i) \\
         &= \frac{1}{N} \sum_{i=1}^{N} \log Z(\theta) -\frac{1}{N}\sum_{i=1}^{N} \left[ \sum_{t=1}^{T} \vec{\phi}(s_{t}) \cdot \vec{\theta} -\frac{1}{2}\left(\| \vec{\phi} \|^2 + \| \vec{\theta} \|^2 \right)  \right]\quad \\
         &= \log Z(\theta)-\frac{1}{N} \sum_{i=1}^{N}\sum_{t=1}^{T} \vec{\phi}(s_{t}) \cdot \vec{\theta} -\frac{1}{2}(\| \vec{\phi} \|^2 + \| \vec{\theta} \|^2)  \\
         &= \log Z(\theta) - \mathbb{E}_{\mu_E} \left[ \vec{\phi}(s_{t}) \cdot \vec{\theta} -\frac{1}{2} (\| \vec{\phi} \|^2 + \| \vec{\theta} \|^2) \right] \\
\mathcal{L}(\theta) &= \log Z(\theta) - \mathbb{E}_{\mu_E} \left[ \vec{\phi}(s_{t}) \cdot \vec{\theta} -\frac{1}{2} ( \| \vec{\theta} \|^2) \right]
\end{align}
We provide the gradient of the loss function with respect to $\theta$:
\begin{align}
    (\nabla_{\theta} \mathcal{L}(\theta))_{i} &= \frac{\partial}{\partial \theta_{i}} \log Z(\theta) - \frac{\partial}{\partial \theta_{i}} \mathbb{E}_{\mu_E} \left[ \vec{\phi}(s_{t}) \cdot \vec{\theta} -\frac{1}{2} ( \| \vec{\theta} \|^2) \right]
\end{align}
We look at the first term of the gradient:
\begin{align}
    &= \frac{\partial}{\partial \theta_i} \log Z(\theta) \\
    &= \frac{1}{Z} \frac{\partial}{\partial \theta_i} \sum_{\tau} e^{R(\tau)} \\
    &= \frac{1}{Z} \sum_{\tau} \frac{\partial e^{R(\tau)}}{\partial R(\tau)} \cdot \frac{\partial R(\tau)}{\partial \theta_i} \\
    &= \frac{1}{Z} \sum_{\tau} e^{R(\tau)} \cdot \frac{\partial R(\tau)}{\partial \theta_i} \\
    &= \frac{1}{Z} \sum_{\tau} e^{R(\tau)} \cdot \frac{\partial}{\partial \theta_i} \sum_{t=1}^{T} \left[ \vec{\phi}(s_{t}) \cdot \vec{\theta} -\frac{1}{2} (\| \vec{\phi} \|^2 + \| \vec{\theta} \|^2) \right] \\
    &= \frac{1}{Z} \sum_{\tau} e^{R(\tau)} \sum_{t=1}^{T} \left[ \vec{\phi}(s_t) - \vec{\theta} \right] \\
    &= \mathbb{E}_{\mu_\pi} \left[ \sum_{t=1}^{T} \vec{\phi}(s_t) - \vec{\theta} \right]
\end{align}
We then take the second term of the gradient in the following way:
\begin{align}
    &= \frac{\partial}{\partial \theta_{i}} -\mathbb{E}_{\mu_E} \left[ \vec{\phi}(s_{t}) \cdot \vec{\theta} -\frac{1}{2} ( \| \vec{\theta} \|^2) \right] \\
    &= -\mathbb{E}_{\mu_E} \left[ \vec{\phi}(s_{t}) - \vec{\theta} \right]
\end{align}
This gives us the following gradient of the loss function with respect to $\theta$ using our Euclidean Distance metric for the
reward function:
\begin{align}
  (\nabla_{\theta} \mathcal{L}(\theta))_i &= \mathbb{E}_{\mu_\pi}\left[ \vec{\phi}(s_t) - \vec{\theta} \right] - \mathbb{E}_{\mu_E} \left[ \vec{\phi}(s_t) - \vec{\theta} \right]
\end{align}

\subsection{Hyperbolic Distance}
\label{app:hyperbolic}
The reward metric is still $R(s_{t})=-\frac{1}{2}d^2$. We use the Lorentzian model where $\mathbb{H}^d$ is represented as $\mathbb{R}^{d+1}$; we define the Minkowski inner product
on $\mathbb{R}^{d+1}$ as $ \langle \vec\theta, \vec\phi \rangle = -\theta_0 \phi_0 + \sum_{n=1}^d \theta_n \phi_n $, and we define the distance as follows: $d=\operatorname{arcosh}(-\langle  \vec{\theta}, \vec{\phi}(s_{t}) \rangle)$,
and therefore $R(s_{t}) = -\frac{1}{2}[\operatorname{arcosh}(-\langle  \vec{\theta}, \vec{\phi}(s_{t}) \rangle)]^2$.

\begin{align}
\mathcal{L}(\theta) &= -\frac{1}{N} \log{P(\mathcal{D}|\theta)} \\
         &= -\frac{1}{N} \log \prod_{i=1}^{N} P(\tau_i|\theta) \\
         &= -\frac{1}{N} \sum_{i=1}^{N} \log P(\tau_i|\theta) \\
         &= -\frac{1}{N} \sum_{i=1}^{N} \log \frac{e^{R(\tau_i)}}{Z(\theta)} \\
         &= -\frac{1}{N} \sum_{i=1}^{N} \log( e^{R(\tau_i)}) + \frac{1}{N} \sum_{i=1}^{N} \log Z(\theta) \\
         &= \frac{1}{N} \sum_{i=1}^{N}\log Z(\theta) -\frac{1}{N}\sum_{i=1}^{N} R(\tau_i) \\
         &= \log Z(\theta) - \frac{1}{N} \sum_{i=1}^{N} \sum_{t=1}^{T} -\frac{1}{2} [\operatorname{arcosh}(\theta_0 \phi_0 - \sum_{n= 1}^{d} \theta_n \phi_n(s_{t}))]^2 \\
\mathcal{L}(\theta) &= \log Z(\theta) - \mathbb{E}_{\mu_E} R(\tau_i)
\end{align}

The gradient of the loss can be written as follows:

\begin{align}
\mathcal{L}(\theta) &= \frac{\partial}{\partial \theta_{i}} \log Z(\theta) - \frac{\partial}{\partial \theta_{i}} \mathbb{E}_{\mu_E} R(\tau_i) \\
\end{align}

We address the first term of the gradient below, first with respect to $\theta_0$ then with respect to $\theta_n$ for $n=1,...,d$:

\begin{align}
    &= \frac{\partial}{\partial \theta_i} \log Z(\theta) \\
    &= \frac{1}{Z} \frac{\partial}{\partial \theta_i} \sum_{\tau} e^{R(\tau)} \\
    &= \frac{1}{Z} \frac{\partial}{\partial \theta_i} \sum_{\tau} e^{\sum_{t=1}^{T} d^2} \\
    &= \frac{1}{Z} \frac{\partial}{\partial \theta_i} \sum_{\tau} e^{\sum_{t=1}^{T} -\frac{1}{2} d^2} \\
    & \intertext{if $x = -(\theta_0 \phi_0 - \sum_{n= 1}^{d} \theta_n \phi_n(s_{t}))$ and $d=\operatorname{arcosh}(x)$} \\
    &= \frac{1}{Z} \sum_{\tau} \frac{\partial e^{R(\tau)}}{\partial \frac{-1}{2}d^2} \cdot \frac{\partial -\frac{1}{2}d^2}{\partial d} \cdot \frac{\partial d}{\partial x} \cdot \frac{\partial x}{\partial \theta_i}  \\
& \intertext{We use the Hyperbolic identity $\cosh^2{d} - 1 = \sinh^2{d}$ and therefore $\sinh(d) = \sqrt{\cosh^2(d) - 1}$} \\
\frac{\partial}{\partial \theta_0} \log Z(\theta) &= \frac{1}{Z}\sum_{i=1}^{N} \sum_{t=0}^{T} \frac{-d e^{-\frac{1}{2}d^2} \phi_0}{\sinh(d)}\\
\frac{\partial}{\partial \theta_n} \log Z(\theta) &= \frac{1}{Z}\sum_{i=1}^{N} \sum_{t=0}^{T} \frac{d e^{-\frac{1}{2}d^2} \phi_n}{\sinh(d)}
\end{align}

The second term of the gradient can be written as follows:

\begin{align}
 &= \frac{\partial}{\partial \theta_i}-\mathbb{E}_{\mu_E} R(\tau_i) \\
 &= -\frac{1}{N}\sum_{i=1}^{N} \sum_{t=0}^{T}\frac{\partial}{\partial \theta_i}R(s_t) \\
 &= -\frac{1}{N}\sum_{i=1}^{N} \sum_{t=0}^{T}\frac{\partial -\frac{1}{2}d^2}{\partial d} \cdot \frac{\partial d}{\partial x} \cdot \frac{\partial x}{\partial \theta_i} \\
 \frac{\partial}{\partial \theta_0} -\mathbb{E}_{\mu_E} R(\tau_i) &=  -\frac{1}{N}\sum_{i=1}^{N} \sum_{t=0}^{T}\frac{-d\phi_0}{\sinh(d)}\\
 \frac{\partial}{\partial \theta_n} -\mathbb{E}_{\mu_E} R(\tau_i) &= \frac{1}{N}\sum_{i=1}^{N} \sum_{t=0}^{T}\frac{-d\phi_n}{\sinh(d)}\\
\end{align}

We combine the first and second derivative terms:
\begin{align}
\partial_{\theta_i}\left(\log Z(\theta)-\mathbb{E}_{\mu_E} R(\tau_i)\right) &=
\left[
\begin{aligned}
\frac{\partial}{\partial \theta_0} &: \frac{1}{Z}\sum_{i=1}^{N} \sum_{t=0}^{T} \frac{-d e^{-\frac{1}{2}d^2} \phi_0}{\sinh(d)} -\frac{1}{N}\sum_{i=1}^{N} \sum_{t=0}^{T}\frac{-d\phi_0}{\sinh(d)}\\[6pt]
\frac{\partial}{\partial \theta_n} &: \frac{1}{Z}\sum_{i=1}^{N} \sum_{t=0}^{T} \frac{d e^{-\frac{1}{2}d^2} \phi_n}{\sinh(d)} + \frac{1}{N}\sum_{i=1}^{N} \sum_{t=0}^{T}\frac{-d\phi_n}{\sinh(d)}
\end{aligned}
\right],
\label{eq:raw-combo-hyp}
\end{align}
where $\partial_{\theta_i}\left(\log Z(\theta) - \mathbb{E}_{\mu_E}R(\tau_i)\right)$ denotes the naive
(Euclidean-sense) partial derivative of the loss. Simplifying each row of Equation~\ref{eq:raw-combo-hyp} gives
\begin{align}
\partial_{\theta_i}\left(\log Z(\theta)-\mathbb{E}_{\mu_E} R(\tau_i)\right) =
\left[
\begin{aligned}
\frac{\partial}{\partial \theta_0} &: -\mathbb{E}_{\mu_\pi}\frac{d\phi_0}{\sinh(d)} + \mathbb{E}_{\mu_E}\frac{d\phi_0}{\sinh(d)}\\[6pt]
\frac{\partial}{\partial \theta_n} &: \mathbb{E}_{\mu_\pi}\frac{d\phi_n}{\sinh(d)} - \mathbb{E}_{\mu_E}\frac{d\phi_n}{\sinh(d)}
\end{aligned}
\right].
\label{eq:naive-grad-hyp}
\end{align}
Note the $\theta_0$ row is the \emph{negative} of the $\theta_n$ row's sign pattern ($-\mu_\pi + \mu_E$ versus
$\mu_\pi - \mu_E$): this is a direct consequence of the Minkowski signature, not an inconsistency -- $\partial
x/\partial\theta_0 = \phi_0$ while $\partial x/\partial\theta_n = -\phi_n$ (Section~\ref{app:hyperbolic}, above),
so the chain rule contributes an extra sign flip for the time-like coordinate that it does not for the
space-like ones.

Because the ambient space carries the Minkowski metric $g=\mathrm{diag}(-1,1,\dots,1)$ rather than the Euclidean
one, the gradient actually used for Riemannian optimization (Section~\ref{sec:exp-map}, Appendix~\ref{app:riemannian})
is not the naive gradient of Equation~\ref{eq:naive-grad-hyp} directly, but its image under $g^{-1}=g$, i.e.\ the
same vector with its time-like ($\theta_0$) component negated. This is a separate step, applied \emph{after} the
derivative itself is fully computed -- it is the same step \citet{praturu2025distance} make explicit when
constructing their analogous ``unconstrained gradient'' $\nabla_u^{d+1}\mathcal{L}_H =
\left(-\partial\mathcal{L}_H/\partial u_0,\ \partial\mathcal{L}_H/\partial u_1,\ \dots\right)$ (their Eq.~2.14) only
after computing $\partial\mathcal{L}_H/\partial u_0$ by ordinary calculus (their Eq.~2.16), noting that ``the
minus sign on the $0$th component of the derivative arises from the signature of the Minkowski inner product.''
Negating the $\theta_0$ row of Equation~\ref{eq:naive-grad-hyp} -- i.e.\ flipping the sign of every term in that
row, turning $-\mu_\pi+\mu_E$ into $\mu_\pi-\mu_E$ -- gives the ambient gradient used throughout the rest of this
paper:
\begin{align}
\nabla^{d+1}_{\theta} &=
\left[
\begin{aligned}
\frac{\partial}{\partial \theta_0} &: \mathbb{E}_{\mu_\pi}\frac{d\phi_0}{\sinh(d)} - \mathbb{E}_{\mu_E}\frac{d\phi_0}{\sinh(d)}\\[6pt]
\frac{\partial}{\partial \theta_n} &: \mathbb{E}_{\mu_\pi}\frac{d\phi_n}{\sinh(d)} - \mathbb{E}_{\mu_E}\frac{d\phi_n}{\sinh(d)}
\end{aligned}
\right] \\
\label{eq:hyp-grad}
\end{align}
which is the exact gradient for the Hyperbolic reward (Section~\ref{sec:hyperbolic}). In practice, $\sinh(d)$
grows exponentially and $1/\sinh(d)$ collapses toward zero for states far from $\theta$, so, following
\citet{praturu2025distance} who report the same difficulty for the analogous term in their model, we drop the
$1/\sinh(d)$ factor and retain only the linear $d$ dependence:
\begin{equation}
    \nabla_\theta \mathcal{L} \approx
    \begin{aligned}[t]
        \frac{\partial}{\partial \theta_0} &: \; \mathbb{E}_{\mu_\pi}\!\left[d\,\phi_0\right] - \mathbb{E}_{\mu_E}\!\left[d\,\phi_0\right] \\[4pt]
        \frac{\partial}{\partial \theta_n} &: \; \mathbb{E}_{\mu_\pi}\!\left[d\,\phi_n\right] - \mathbb{E}_{\mu_E}\!\left[d\,\phi_n\right]
    \end{aligned}
    \label{eq:approx-grad}
\end{equation}
All Hyperbolic results reported in this paper use this practical gradient.

This practical gradient is the starting point for
the tangent-space projection of Section~\ref{sec:exp-map}, below.

\subsection{Riemannian Optimization: Tangent-Space Projection and the Exponential Map}
\label{sec:exp-map}
\label{app:riemannian}

Equation~\ref{eq:hyp-grad} (or its practical approximation, Equation~\ref{eq:approx-grad}) is an \emph{ambient}
gradient in $\mathbb{R}^{d+1}$: it does not account for the
constraint $\langle \vec\theta, \vec\theta \rangle_{\mathcal M} = -1$ that keeps $\theta$ on the hyperboloid. Taking
$\theta \leftarrow \theta - \eta \nabla_\theta \mathcal{L}$ directly would therefore move $\theta$ off the manifold.
We instead perform Riemannian gradient descent \citep{nickel2018learning, wilson2018gradient, praturu2025distance}: project the
ambient gradient onto the tangent space at $\theta$, then take a step along the geodesic in that direction via
the exponential map, so that every update remains exactly on $\mathbb{H}^d$.

\textbf{Tangent-space projection.} The tangent space at $\theta$ is $T_\theta \mathbb{H}^d = \{ v \in
\mathbb{R}^{d+1} : \langle \vec\theta, v \rangle_{\mathcal M} = 0 \}$. The Riemannian gradient is obtained from
the ambient gradient by adding back its component along $\theta$, using the Minkowski inner product in place of
the Euclidean one:
\begin{equation}
    \nabla^{\mathcal H}_\theta \mathcal{L} = \nabla_\theta \mathcal{L} + \left\langle \vec\theta, \nabla_\theta \mathcal{L} \right\rangle_{\mathcal M} \vec\theta.
    \label{eq:riem-grad}
\end{equation}

\textbf{Exponential map.} For a tangent vector $y \in T_\theta \mathbb{H}^d$, the exponential map
\begin{equation}
    \mathrm{Exp}_\theta(y) = \cosh\!\left( \sqrt{\langle y, y \rangle_{\mathcal M}} \right) \vec\theta
    + \sinh\!\left( \sqrt{\langle y, y \rangle_{\mathcal M}} \right) \frac{y}{\sqrt{\langle y, y \rangle_{\mathcal M}}}
    \label{eq:exp-map}
\end{equation}
takes a step of geodesic length $\sqrt{\langle y, y \rangle_{\mathcal M}}$ along $\mathbb{H}^d$ in the direction of
$y$. We replace the Euclidean update rule with
\begin{equation}
    \theta \leftarrow \mathrm{Exp}_\theta\left( -\eta \, \nabla^{\mathcal H}_\theta \mathcal{L} \right),
    \label{eq:exp-step}
\end{equation}
for learning rate $\eta$, i.e.\ $y = -\eta \nabla^{\mathcal H}_\theta \mathcal{L}$ in Equation~\ref{eq:exp-map}.

For the radius-$R$ hyperboloid (Section~\ref{sec:curvature}), the Riemannian gradient correction
(Equation~\ref{eq:riem-grad}) picks up a compensating factor of $1/R^2$,
\begin{equation}
    \nabla^{\mathcal H}_\theta \mathcal{L} = \nabla_\theta \mathcal{L} + \frac{1}{R^2}\left\langle \vec\theta, \nabla_\theta \mathcal{L} \right\rangle_{\mathcal M} \vec\theta,
\end{equation}
since $\langle \vec\theta, \vec\theta \rangle_{\mathcal M} = -R^2$ rather than $-1$, and the exponential map
(Equation~\ref{eq:exp-map}) becomes
\begin{equation}
    \mathrm{Exp}_\theta(y) = \cosh\!\left( \frac{\|y\|_{\mathcal M}}{R} \right) \vec\theta
    + R \sinh\!\left( \frac{\|y\|_{\mathcal M}}{R} \right) \frac{y}{\|y\|_{\mathcal M}}, \qquad
    \|y\|_{\mathcal M} \equiv \sqrt{\langle y, y \rangle_{\mathcal M}},
\end{equation}
where $R=1$ recovers Equations~\ref{eq:riem-grad}--\ref{eq:exp-map}.

\subsection{Numerical Stability of the Exponential-Map Step}
\label{app:step-clipping}

The exponential-map update (Equation~\ref{eq:exp-step}) is only numerically safe for moderate step sizes. Two
failure modes appear in practice, both traceable to the same cause: as $\theta$ drifts away from the origin of
the hyperboloid over the course of training, curvature makes a fixed step size $\eta$ eventually produce an
unstable update, regardless of how $\theta$ got there. (We verified that this instability is intrinsic to the fixed
step size rather than a sampling-noise artifact from the policy rollouts used to estimate $\mu_\pi$: increasing
the number of rollouts by an order of magnitude did not change when the instability appeared.) Writing
$y = -\eta \nabla^{\mathcal H}_\theta \mathcal{L}$ for the proposed tangent step before it is applied:
\begin{enumerate}
    \item \textbf{Loss of precision in $\langle y, y \rangle_{\mathcal M}$.} This Minkowski norm is computed as a
    difference of two large, nearly equal floating-point sums once $y$'s components exceed roughly $10^{8}$, so it
    loses precision to catastrophic cancellation and can come out negative -- i.e.\ not a valid tangent vector --
    even when $y$ is mathematically valid.
    \item \textbf{Overflow in $\cosh/\sinh$.} Even when $\langle y, y \rangle_{\mathcal M}$ is computed correctly,
    $\cosh(\sqrt{\langle y,y\rangle_{\mathcal M}})$ and $\sinh(\sqrt{\langle y,y\rangle_{\mathcal M}})$ overflow
    double-precision floating point once $\sqrt{\langle y,y\rangle_{\mathcal M}}$ is roughly $709$ or larger, producing
    \texttt{inf}/\texttt{NaN} values that propagate into the softmax policy computed on the next iteration.
\end{enumerate}
We address both failure modes at their common source by capping the geodesic step length before either
$\langle y, y \rangle_{\mathcal M}$ or the exponential map is evaluated: given a maximum step norm $S_{\max}$, if
$\eta \left\| \nabla^{\mathcal H}_\theta \mathcal{L} \right\|_2 > S_{\max}$ (using the ordinary Euclidean norm,
which is cheap to compute and tracks the eventual geodesic step length monotonically), we rescale
\begin{equation}
    \eta \leftarrow \eta \cdot \frac{S_{\max}}{\eta \left\| \nabla^{\mathcal H}_\theta \mathcal{L} \right\|_2}
\end{equation}
before taking the step -- i.e.\ the learning rate is only ever reduced, never increased, and the rescaling
happens before any Hyperbolic quantity is computed. We use $S_{\max} = 5$ geodesic units throughout.

\subsection{Mixed Second Difference}
\label{app:mixed-second-difference}
To summarize non-additivity in the reward surface as a single number per feature pair (Section~\ref{sec:feature-coupling}),
we compute the mixed second difference of the reward surface, evaluated at two corner values
$\text{hi}=0.9$ and $\text{lo}=0.1$ of each feature:
\begin{equation}
    \Delta^2 R = R(\text{hi},\text{hi}) - R(\text{hi},\text{lo}) - R(\text{lo},\text{hi}) + R(\text{lo},\text{lo}).
\end{equation}
This is the discrete analogue of the mixed partial derivative $\partial^2 R/\partial\phi_i\partial\phi_j$,
and like the residual of the two-way ANOVA-style fit (Section~\ref{sec:feature-coupling}), it is exactly
zero whenever $R$ is additively separable in $\phi_i$ and $\phi_j$. Throughout the rest of the paper we
refer to this quantity as the conjunctive effect.

\subsection{Museum Exhibit Reward Magnitude}
\label{app:exhibit-reward-magnitude}
\begin{figure}[htb]
\centering
\includegraphics[width=0.75\linewidth]{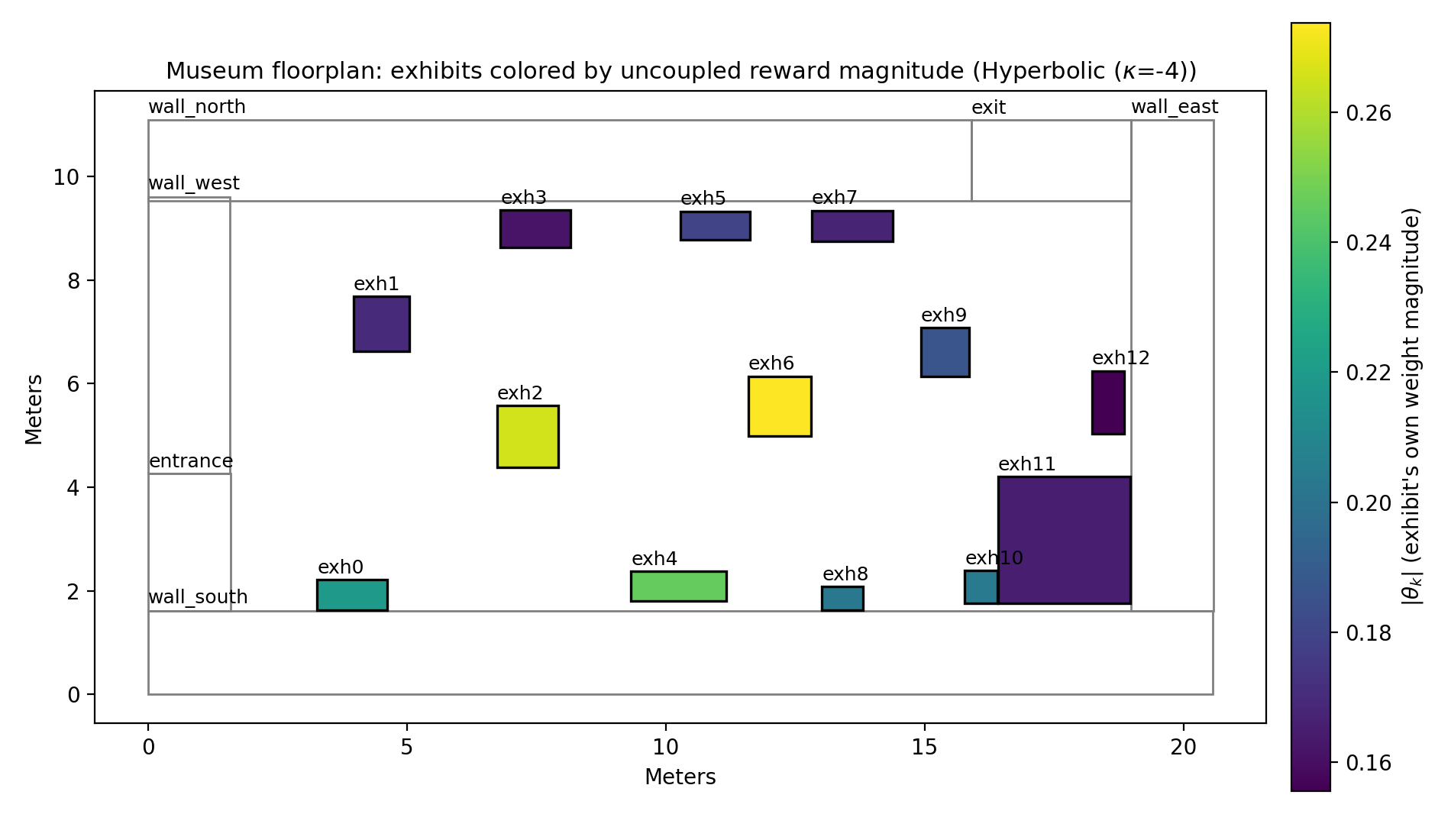}
\caption{Museum floorplan with each exhibit colored by its own (uncoupled) fitted reward weight
magnitude $|\theta_k|$, Hyperbolic ($\kappa$=-4).}
\label{fig:Exhibit Reward Magnitude}
\end{figure}

Figure~\ref{fig:Exhibit Reward Magnitude} shows the museum floorplan with each exhibit colored by
$|\theta_k|$, the magnitude of that exhibit's own fitted weight (Hyperbolic, $\kappa$=-4), i.e. how
much reward that exhibit contributes on its own, uncoupled from any other exhibit.

\FloatBarrier

\subsection{Permutation (Mantel) Test for Reward-Coupling Alignment}
\label{app:mantel-test}
The exhibit-pair alignment test in Section~\ref{sec:feature-coupling} correlates $\mathrm{co\_attendance}(i,j)$,
the real behavioral correlation between how closely visitors approach exhibits $i$ and $j$ in the same
trajectory, against each geometry's fitted conjunctive effect for that pair. Because the 78 exhibit pairs
are built from only 13 exhibits (each exhibit appears in 12 of the 78 pairs), the pairs are not
independent samples, so the naive $p$-values from that correlation are likely optimistic. We correct for
this with a Mantel-style permutation test: we permute the 13 exhibit labels jointly across rows and
columns of the $\mathrm{co\_attendance}$ matrix (holding the conjunctive-effect matrix fixed), recompute the
correlation, and repeat for 10,000 shuffles to build an empirical null distribution. This preserves the
pairwise dependency structure among the 78 correlations (since permuting exhibit labels permutes all 12
pairs involving a given exhibit together) and is the standard way to test correlations computed over
pairs drawn from a smaller set of underlying units.

The resulting correlations and permutation-corrected $p$-values are reported in Table~\ref{tab:mantel}
below, corresponding to the scatter plots in Figure~\ref{fig:CoAttendance Scatter} in the main text
(Section~\ref{sec:feature-coupling}). As a further check, we also residualize both $\mathrm{co\_attendance}$
and the fitted conjunctive effect against physical exhibit-to-exhibit distance before correlating (partial
correlation columns), to rule out spatial proximity alone as a confound; this leaves the pattern
essentially unchanged for every geometry (e.g., for Hyperbolic ($\kappa$=-4), partial $r=+0.539$ vs.\ raw
$r=+0.516$), with the same three curvatures remaining significant under permutation correction. We also
report the naive (non-permutation) $p$-value alongside the corrected one to make explicit how much the
non-independence of the 78 pairs inflates apparent significance: e.g., Hyperbolic ($\kappa$=-4)'s naive
$p\approx1.3\times10^{-6}$ becomes $p=0.0026$ once corrected, still significant but three orders of
magnitude less extreme.

\begin{table}[htb]
\centering
\caption{Alignment between real exhibit co-attendance and each geometry's fitted conjunctive effect: raw
correlation with its naive (non-permutation) and permutation-corrected $p$-values (10,000 shuffles of
exhibit labels), and the same three quantities after residualizing both variables against physical
exhibit-to-exhibit distance (partial correlation).}
\label{tab:mantel}
\begin{tabular}{lcccccc}
\toprule
& \multicolumn{3}{c}{Raw} & \multicolumn{3}{c}{Partial (distance-controlled)} \\
\cmidrule(lr){2-4} \cmidrule(lr){5-7}
Geometry & $r$ & naive $p$ & perm.\ $p$ & $r$ & naive $p$ & perm.\ $p$ \\
\midrule
Classic & -0.070 & 0.54 & 0.48 & -0.048 & 0.68 & 0.65 \\
Euclidean & +0.075 & 0.51 & 0.53 & +0.058 & 0.62 & 0.63 \\
Hyperbolic ($\kappa$=-25) & +0.477 & $1.0\times10^{-5}$ & 0.0021 & +0.507 & $2.2\times10^{-6}$ & 0.0011 \\
Hyperbolic ($\kappa$=-4) & +0.516 & $1.3\times10^{-6}$ & 0.0026 & +0.539 & $3.5\times10^{-7}$ & 0.0019 \\
Hyperbolic ($\kappa$=-1) & +0.419 & $1.3\times10^{-4}$ & 0.0034 & +0.421 & $1.2\times10^{-4}$ & 0.0035 \\
DeepIRL & +0.205 & 0.071 & 0.084 & +0.228 & 0.045 & 0.056 \\
\bottomrule
\end{tabular}
\end{table}
\FloatBarrier

This permutation test only corrects for the non-independence among the 78 \emph{exhibit-pair} samples;
it does not address a separate form of non-independence at the \emph{trajectory} level (each visitor
trajectory contributes to the $\mathrm{co\_attendance}$ value of multiple exhibit pairs), which is not
addressed here.

\raggedbottom
\makeatletter
\setlength{\@fptop}{0pt}
\makeatother

\end{document}